\documentclass[conference]{IEEEtran}
\IEEEoverridecommandlockouts
\usepackage{cite}
\usepackage{amsmath,amssymb,amsfonts}
\usepackage{algorithmic}
\usepackage{graphicx}
\usepackage{textcomp}
\usepackage{xcolor}
\usepackage{algorithm}
\usepackage{array}
\usepackage[caption=false,font=normalsize,labelfont=sf,textfont=sf]{subfig}
\usepackage{booktabs}
\usepackage{stfloats}
\usepackage{url}
\usepackage{verbatim}
\usepackage{graphicx}
\usepackage{enumitem}
\usepackage{multirow}
\usepackage[graphicx]{realboxes}
\usepackage{float}  
\usepackage{bbding}
\usepackage{hyperref}
\usepackage{tcolorbox}
\usepackage{xspace}
\usepackage[normalem]{ulem}
\useunder{\uline}{\ul}{}
\newtheorem{theorem}{Theorem}
\newtheorem{lemma}{Lemma}

\def\BibTeX{{\rm B\kern-.05em{\sc i\kern-.025em b}\kern-.08em
    T\kern-.1667em\lower.7ex\hbox{E}\kern-.125emX}}
\begin{document}

\title{PRUNE: A Patching Based Repair Framework for Certifiable Unlearning of Neural Networks
}

\author{
Xuran Li$^{1, 3}$, Jingyi Wang$^{1, 3 *}$, Xiaohan Yuan$^{1}$, Peixin Zhang$^{2}$ \\
$^{1}$ Zhejiang University, Hangzhou, China 
$^{2}$ Singapore Management University, Singapore \\
$^{3}$ Huzhou Institute of Industrial Control Technology, Huzhou, China\\
\texttt{\{xuranli1005, wangjyee, xiaohanyuan\}@zju.edu.cn}, \texttt{pxzhang94@gmail.com}
}
\maketitle

\begin{abstract}
It is often desirable to remove (a.k.a. unlearn) specific training data from a neural network model, especially to uphold the data holder’s \emph{right to be forgotten} as mandated by recent regulations. Beyond privacy concerns, unlearning is also increasingly relevant in industrial settings, such as in adaptive control systems or predictive maintenance pipelines, where removing outdated or faulty sensor data is critical for system safety and performance. Existing unlearning methods involve training alternative models with remaining data, which may be costly. In this work, we provide a new angle and propose a novel unlearning approach by imposing carefully crafted \emph{‘patch’} on the original neural network to achieve targeted \emph{‘forgetting’} of the requested data to delete. Specifically, inspired by the research line of \emph{neural network repair}, we propose to strategically seek a lightweight minimum ‘patch’ for unlearning a given data point with certifiable guarantee. Extensive experiments on multiple categorical datasets demonstrates our approach's effectiveness, achieving measurable unlearning while preserving the model's performance.
\end{abstract}


\section{Introduction}

In the era of data-driven machine learning, legal frameworks such as the GDPR \cite{regulation2018general} grant users the \emph{right to be forgotten}, allowing them to request removal of their data from trained models \cite{ginart2019making}. 
To support the request of data erasure (or removal), the study on \textit{machine unlearning} \cite{bourtoule2021machine,zhou2025decoupled,yan2022arcane,foster2024fast} has emerged. Arguably, the most intuitive way to unlearn is to retrain an alternative model $\textsc{M}_r$ after removing the data to be withdrawn from the training set. We use $\textsc{M}_r$ to denote the retrained model without the data to remove and $\textsc{M}_U$ the model obtained after different kinds of unlearning methods consistently. Existing unlearning methods are mostly measured by the distance between the unlearned model $\textsc{M}_U$ and $\textsc{M}_r$, and can be roughly divided into two categories: \textit{exact unlearning} and \textit{approximate unlearning}. The general idea of exact unlearning is to retrain a model (without the data to remove) using different speedup approaches \cite{bourtoule2021machine,hu2024separate}. They often perform additional operations during the model training phase to reduce the cost of retraining by, for example, either slicing the data \cite{bourtoule2021machine} or segmenting the training \cite{yan2022arcane}.
Approximate unlearning \cite{zhang2022prompt,chien2023efficient,foster2024fast,ginart2019making}, on the other hand, aims to approximate the performance of $\textsc{M}_r$ by modifying the model parameters to save the retraining cost. For example, the influence function is used to estimate the impact of data withdrawal on the model, and the model parameters are updated on this basis \cite{wu2022puma,wu2023gif}.

In this work, we propose a \textbf{P}atching based \textbf{R}epair framework for certifiable \textbf{UN}learning (\textbf{PRUNE}). It draws an analogy between data removal in unlearning and error correction in \emph{neural network repair} \cite{dong2021towards,ma2024vere,sohn2022arachne}. By doing so, we formulate the unlearning problem as a neural network repair problem, where a similar operation with the opposite objective function can be performed to satisfy the needs of both the data holder and the model owner. Specifically, we first propose to carefully craft a minimum `patch' network for unlearning a targeted given data point by redirecting the model's prediction elsewhere in a certifiable way. This approach is particularly relevant for scenarios where real-world systems, such as industrial automation or adaptive control pipelines. They must quickly forget outdated or incorrect sensor records without halting operation or undergoing costly retraining.
We extensively evaluated the effectiveness of our approach on multiple categorical datasets. The results show that our approach can achieve easily measurable unlearning
while retaining the model's original performance on the remaining data.
Besides, our approach is competitive in terms of efficiency and memory consumption in comparison with various baseline unlearning methods.

In summary, we make the following main contributions:

\begin{itemize}
  \item We propose a new approach that bridges machine unlearning and neural network repair. By applying a carefully designed patch network, we directly falsify the model’s prediction on the withdrawn data.
  \item To preserve the model’s performance on the remaining data, we ensure that the patch network is both minimal and localized. 
  \item We evaluate our approach on multiple categorical datasets\cite{anony2023code}, showing effective forgetting with competitive efficiency and resource usage.
\end{itemize}

\section{Methodology}

\subsection{Problem Definition}
\label{UR}
In this paper, we focus on how to unlearn data on Deep Neural Networks (DNNs) used for classification tasks. This is a very common task in many real-world scenarios. Formally, let $\mathcal{D}=\left\{  x_i,y_i  \right\} _{i=1,...,n}$ be the dataset containing data points $x_i$ and corresponding labels $y_i$. Given a DNN model $M:\mathcal{X}\rightarrow\mathcal{Y}$, $\mathcal{X}$ is the input domain and $\mathcal{Y}=\left\{ 1,2,...,L \right\}$ is the set of category labels. $M_\mathcal{D}$ is a DNN model trained on the dataset $\mathcal{D}$. It makes judgment about the label of $x_i:\mathrm{arg}\max_{l \in \mathcal{Y}}  M_{\mathcal{D}}^{l}\left( x_i \right)$, where $M_{\mathcal{D}}^{l}(x)$ denotes the output probability that model $M_\mathcal{D}$ considers the label of $x$ to be $l$. When there are requests for erasure, $\mathcal{D}_U=\left\{ x_u,y_u \right\} _{u=1,...,r}$ denotes the set of data points to be unlearned and $\mathcal{D}_U\subset\mathcal{D}$. $\mathcal{D}_R=\mathcal{D}/\mathcal{D}_U$ is the set of remaining data points. $M_U$ is the model after the execution of the unlearning algorithm. 

From the data holder's perspective, unlearning is considered to be effective intuitively on $x_u$ iff $\mathrm{arg}\max_{l \in \mathcal{Y}}  M_{U}^{l}\left( x_u \right) \ne y_u$, i.e., the unlearned model can no longer makes the correct judgement on his/her data. 
This objective can be formalized as for any $x_u \in \mathcal{D}_U$,
\begin{equation}\label{eq3}Obj^u_1:
\exists l\ne y_u,M_{U}^{l}\left( x_u \right) -M_{U}^{y_u}\left( x_u \right) >0
\end{equation}
Note that $Obj^u_1$ and $Obj^r_1$ are two exactly mirrored operation.

Besides the unlearning objective in Equation (\ref{eq3}), similar to the repair task, we also have to pay attention to the overall performance of the model on the remaining data. 
The ideal situation is that the model produces no change in label judgments on $\mathcal{D}_R$. Thus, the other objective in Equation (\ref{eq4}) is to make the performance change on the remaining data as small as possible. It is formulated as
\begin{equation}\label{eq4}
Obj^u_2: min\quad \frac{1}{|\mathcal{D}_R|}\sum_{i=1}^{|\mathcal{D}_R|}{|} \mathrm{arg}\max _{l \in \mathcal{Y}} M_{U}^{l}\left( x_i \right) -\mathrm{arg}\max _{l \in \mathcal{Y}} M_{\mathcal{D}}^{l}\left( x_i \right)  |
\end{equation}

When both objectives are achieved, the unlearning algorithm can be considered as meeting both the needs of data holders and the interests of the model owner. Our approach is designed precisely in accordance with these objectives.

\subsection{The PRUNE Framework}

\begin{figure*}[h]
  \centering
  \includegraphics[width=0.9\linewidth]{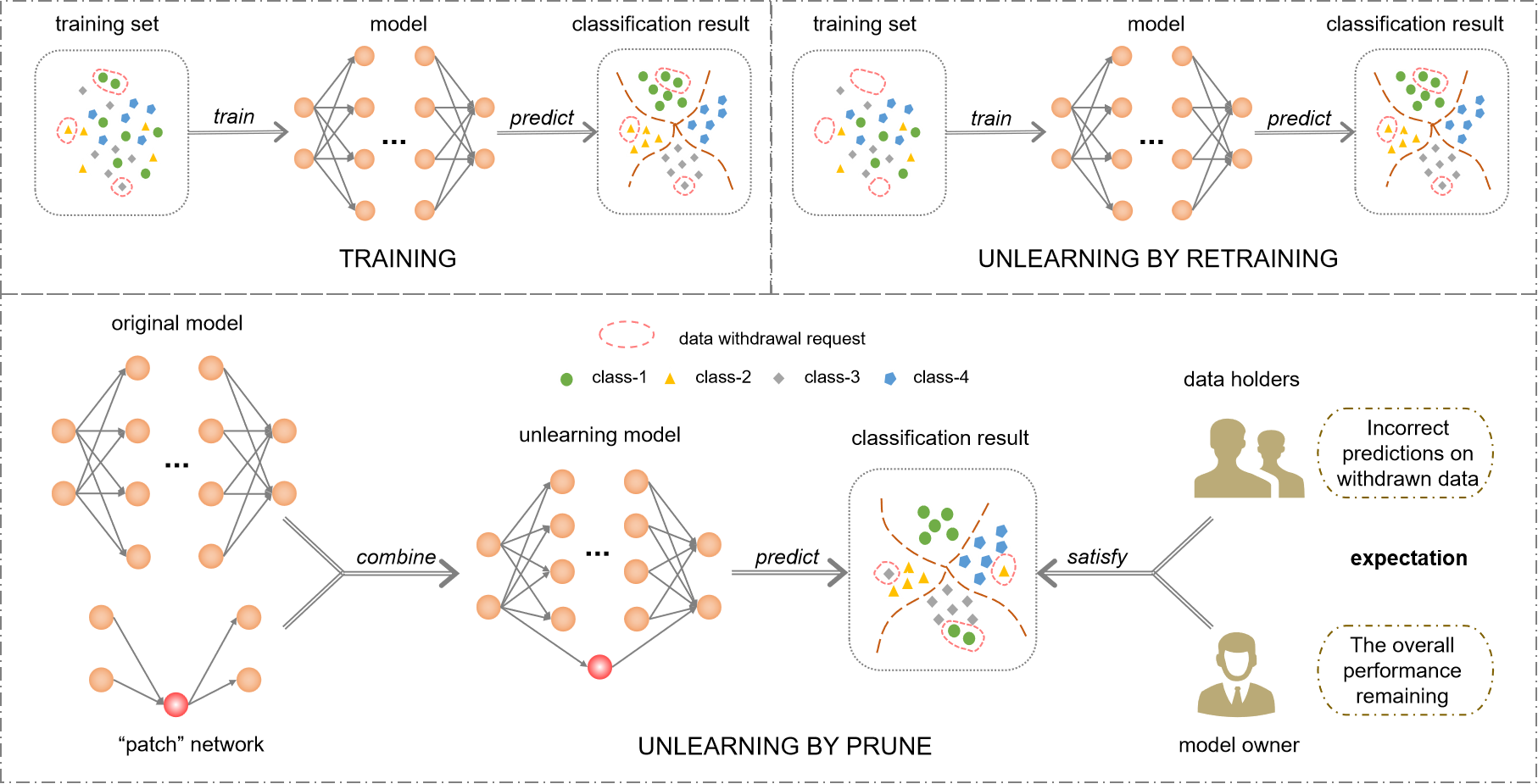}
  \caption{The framework of PRUNE. }
  \label{sys-overview}
\end{figure*}

In the following, we present our PRUNE in details. As shown in Figure \ref{sys-overview}, the key idea of PRUNE is to generate a targeted ``patch'' network on the original model $M_\mathcal{D}$ training on $\mathcal{D}$ to unlearn the specified data, i.e., redirecting the model's prediction to elsewhere wrong. The patch is \emph{targeted} in the sense that there is a one-to-one mapping from the specified data point to unlearn $x_u$ to the patch network $c(x)$. And the patch network will only be activated when running the model on the specific data point, which means the model's performance will not be affected on any other data than the data to unlearn. In the following, we present the technical details of how we realize the idea of PRUNE and generate the patch network for different unlearning scenarios. 

We denote a DNN model by the concatenation of two sequential parts $M=M_p\oplus M_c$, where $M_p$ is used for feature extraction with operations like convolution and $M_c$ are the fully connected layers. 
In general, our approach is applicable to continuous piecewise linear (CPWL) neural networks, i.e., $M_c$ with Rectified Linear Unit (ReLU) activation function $\mathrm{ReLU} \left( x \right) =\max \left( 0,x \right)$. We do not have requirements on $M_p$. 
The proof details of the theorems are provided in \cite{anony2023code}.

\begin{theorem}
\label{the:main}
    Given a neural network model $M_{\mathcal{D}}=M_p\oplus M_c$ trained on $\mathcal{D}$, and a sample data $x_u$ to unlearn, it is guaranteed that we can construct a patch network $c_S$ for $M_c$ and obtain an unlearned model $M_U=M_p\oplus (M_c+c_S)$ such that:
    1) $ M_U(x_u)\neq M_{\mathcal{D}}(x_u)$; and
    2) for $x \in \mathcal{D}/x_u$, $ M_U(x)= M_{\mathcal{D}}(x)$.  
\end{theorem}

Next, we introduce how the patch network $c_S$ is constructed. In general, a \emph{patch network} consists of two parts: a \emph{confusion sub-network} that affects the output domain (directing the prediction on $x$ elsewhere) and a \emph{support sub-network} that limits the side effect (not affecting the model's prediction on the remaining data). 
The construction has three steps whose details are as follows.

\emph{1) Locating the linear region of $x_u$.} Considering our unlearning goal in Equation \ref{eq3}, we should pay more attention to the correspondence between the input domain and the output domain, while ignoring the model structure change. Since our study object is
CPWL, the linear region where the data point $x_u$ to be unlearned lies can be first computed similarly to \cite{lee2019towards}.

\begin{lemma}
\label{le:fs}
Given a CPWL neural network with neurons $z$, each $z_{j}^{i} \in z$ induces a feasible set $S_{j}^{i}\left( x \right)$ for input $x_u \in \mathcal{X}$. For $\bar{x_u}\in \mathcal{X}$,
\begin{equation}\label{eq5}
S_{j}^{i}\left( x_u \right) =\left\{ \begin{array}{l}
	\left( \nabla _{x_u}z_{j}^{i} \right) ^T\bar{x_u}+z_{j}^{i}-\left( \nabla _{x_u} z_{j}^{i} \right) ^T{x_u} \geq 0,z_{j}^{i}\ge 0\\
	\left( \nabla _{x_u} z_{j}^{i} \right) ^T\bar{x_u}+z_{j}^{i}-\left( \nabla _{x_u} z_{j}^{i} \right) ^T{x_u} \le 0,z_{j}^{i}<0\\
\end{array} \right.  
\end{equation}    
\end{lemma}
$z_{j}^{i}$ denotes the $i$-th neuron (before activation) in the $j$-th hidden layer and $\nabla _{x_u}z_{j}^{i}$ is the sub-gradient calculated by back-propagation. The linear region including $x_u$ is the feasible set $S\left( x_u \right) =\cap _{i,j}S_{j}^{i}\left( x_u \right) $. 

Assume that $S\left( x_u \right) =\left\{ a_ix_u\leq b_i \right\}_{i=1,2,...,N} $ is a linear region calculated by Lemma \ref{le:fs}, which is composed of $N$ inequalities. To achieve the goal in Equation \ref{eq4} (not affecting the remaining data), we further use a support network to restrict the subsequent unlearning effect to occur only on $S(x_u)$ defined by the following lemma \cite{fu2022sound}.

\begin{lemma}
For a neural network using ReLU function as the activation function, the support network on the given feasible set $S(x_u)$ is defined as
\begin{equation}\label{eq6}
n_S\left( x_u,\lambda \right) =\mathrm{ReLU} \left( \sum_i{n\left( b_i-a_ix_u,\lambda \right) -N+1} \right) 
\end{equation}
where $n(x_u,\lambda)=\mathrm{ReLU}(\lambda x_u+1)-\mathrm{ReLU}(\lambda x_u)$ and $\lambda$ is a parameter to control the boundary of $n(x_u)$. 
\end{lemma}

\emph{2) Optimizing the confusion network.} $Obj^u_1$ requires $M_U$ to fail to predict on $x_u$. This is the most basic utility of the confusion network $m(x_u)$, i.e., $M_U\left( x_u \right) =M_p(x_u)\oplus (M_c(x_u) +m\left( x_u \right)) \rightarrow \hat{y}_u\ne y_u$. Meanwhile, $m(x_u)$ is expected to be minimal in the function space to reduce the impact on the model performance. Thus $m(x)$ for unlearning $x_u$ can be optimized based on the following Lemma.
\begin{lemma}
Given a neural network model $M_{\mathcal{D}}=M_p\oplus M_c$, and a sample data $x_u \in \mathcal{D}$ with the feasible set $S(x_u)$, the optimization objectives of the confusion network $m(x)=Cx+d$ can be formalized as
\begin{equation}\label{eq7}
\left\{ \begin{array}{l}
	\min_{C,d}\max_{x\in S(x_u)}|m\left( x \right) |\\
	M_U\left( x \right) =M_p(x)\oplus (M_c(x) +m\left( x \right))\\
	\forall x\in S\left( x_u \right) ,M_{U}^{\hat{y}_u}\left( x \right) -M_{U}^{l}\left( x \right) >0,l\ne \hat{y}_u\in \mathcal{Y}\\
\end{array} \right. 
\end{equation}
where $C$ is a matrix, $d$ is a vector, and $\hat{y}_u$ is the confusing label randomly taken in $\mathcal{Y} \backslash y_u$.
\end{lemma}


\emph{3) Combining into patch network.} For a single data point $x_u$ in $S$ to unlearn, the patch network $c\left( x_u \right) $ can be expressed as 
\begin{equation}\label{eq8}
\begin{split}
c_S\left( x_u \right) &=\mathrm{ReLU}\left( m_S\left( x_u \right) +H\cdot n_S\left( x_u,\lambda \right) -H \right)\\
&-\mathrm{ReLU}\left( -m_S\left( x_u \right) +H\cdot n_S\left( x_u,\lambda \right) -H \right) 
\end{split}
\end{equation}
where $H$ is the upper bound of $m(x)$ obtained in the linear region $S$. 

Based on the above steps, Theorem \ref{the:main} provides ideal and certifiable unlearning guarantee for a given data point by producing a targeted patch. Next, we further illustrate how to efficiently apply PRUNE to handle unlearning of multiple data points.
\subsubsection*{\textbf{Unlearning on Multiple Data Points}}
\label{multipoint}
Note that the main cost of our algorithm is on the optimization of Equation \ref{eq7}. Considering the time complexity, for the dataset $\mathcal{D}_U$ to unlearn, we expect to optimize fewer confusion networks to affect more labels of data points. $m(x)$ is optimized based on a linear region where a single $x_u$ is located, so our intuition is to choose the most representative data points to generate $m(x)$. Here, we use clustering to select representative data points as follows. The data points ${ x_u }_{u=1,\dots,r}$ are grouped into $K$ clusters using K-Means, where each cluster has a centroid $x_c^k$.
Since $ x_{c}^{k} $ has the shortest Euclidean distance from all points in their cluster $\left\{ x_{u}^{k} \right\}_{u=1,...,r'}$, we use $x_{c}^{k}$ as the representative point to generate confusion network $m_k$ according to Equation \ref{eq7}. We feed $\left\{ x_{u}^{k} \right\}_{u=1,...,r'}$ into the temporary network $M_\mathcal{D}+m_k$ to test whether the label of $x_{u}^{k}$ can still be judged correctly. If the label has been misjudged, the corresponding support network $n_{u}^{k}$ is calculated according to Equation \ref{eq6}. If the output of this point has not changed, it will be recorded into the remaining unlearning dataset $\mathcal{D}_{UR}$ for a new round of iteration. After traversing all points in $\mathcal{D}_{U}^{k}$, a confusion network $m_k$ and a series of support networks $\left\{ n_{u}^{k} \right\}_{u=1,...,r'}$ are obtained, so the corresponding patch network $c_k$ can be calculated by
\begin{equation}\label{eq10}
\begin{split}
c_k\left( x,\lambda \right) &=\mathrm{ReLU}\left( m_k\left( x \right) +\max _{1\le u\le r'}\left\{ n_{u}^{k}\left( x,\lambda \right) \right\} \cdot H_u-H_u \right) \\
&-\mathrm{ReLU}\left( -m_k\left( x \right) +\max _{1\le u\le r'}\left\{ n_{u}^{k}\left( x,\lambda \right) \right\} \cdot H_u-H_u \right) 
\end{split}
\end{equation}
where $H_u=\max \left\{ \left| m_k\left( x \right) \right||x\in \cup _{1\le u\le r'}S\left( x_{u}^{k} \right) \right\} $. And when all clusters in $\mathcal{D}_U$ have been optimized $m_k$, tested, generated $\left\{ n_{u}^{k} \right\}_{u=1\to r'}$ and calculated $c_k$, $M_c$ can be updated to $M_c+\left\{ c_k \right\} _{k=1\to K}$. Finaly, we judge whether to end the entire iterative process by unlearning success rate $1-\mathcal{D}_{UR}/\mathcal{D}_U$. When it is higher than the required $\delta$, the algorithm is completed. The overall process of PRUNE for multipoint unlearning is summarized in Algorithm \ref{ourapproach} which is guaranteed to terminate and we thus have the following theorem.

\begin{theorem}
\label{the:multi}
    Given a neural network model $M_{\mathcal{D}}=M_p\oplus M_c$ trained on $\mathcal{D}$, a set of data $\mathcal{D}_U\subset \mathcal{D}$ to unlearn and a desired unlearning degree $\delta$ on $\mathcal{D}_U$, it is guaranteed that we can construct a series of patch network $\{c_k\}_{k=1\to K}$ for $M_c$ and obtain an unlearned model $M_U=M_p\oplus M_c+(\{c_k\}_{k=1\to K})$ such that: for $x \in \mathcal{D}_U$, $Pr(M_U(x)\neq M_{\mathcal{D}}(x))\geq \delta$.
\end{theorem}

\begin{algorithm}[tb]
   \caption{PRUNE-Multipoint}
   \label{ourapproach}
\begin{algorithmic}[1]
   \STATE {\bfseries Input:} $\mathcal{D}_U,M_\mathcal{D}=M_p\oplus M_c$
   \STATE {\bfseries Output:} $M_U$
   \STATE Initialize clusters number $K$
   \STATE $\mathcal{D}_{U}^{k},\left\{ x_{c}^{k},y_{c}^{k} \right\} _{k=1\to K} \gets {\rm KMeans}(\mathcal{D}_U,K)$
   \FOR{$k \gets 1$ to $K$}
       \STATE $\hat{y}_{c}^{k} \gets$ Randomize $\mathcal{Y} \backslash y_{c}^{k}$
       \STATE $m_k \gets$ according to Eq. \ref{eq7}
       \FOR{$(x_{u}^{k},y_{u}^{k}) \in \mathcal{D}_{U}^{k}$}
           \IF{$M_p(x_{u}^{k})\oplus (M_c(x_{u}^{k}) +m_k\left( x_{u}^{k} \right))!=y_{u}^{k}$}
               \STATE $n_{u}^{k} \gets$ according to Eq. \ref{eq6}
           \ELSE
               \STATE $\mathcal{D}_{UR} \gets (x_{u}^{k},y_{u}^{k})$
           \ENDIF
       \ENDFOR
       \STATE $c_k \gets$ according to Eq. \ref{eq10}
   \ENDFOR
   \STATE $M_c \gets M_c+\left\{ c_k \right\} _{k=1\to K}$ 
   \REPEAT 
   \STATE $\mathcal{D}_U \gets \mathcal{D}_{UR}$
   \STATE line4-line17
   \UNTIL{$1-\mathcal{D}_{UR}/\mathcal{D}_U>\delta$}
   \STATE $M_U=M_\mathcal{D}$
\end{algorithmic}
\end{algorithm}
\section{Experiments}
In this section, we describe the experimental setup and conduct extensive experiments aiming to answer the following research questions:
    \begin{itemize}
    \item \textbf{RQ1:} Can our approach forget specific data points while keeping the model performance as constant as possible?
    \item \textbf{RQ2:} How does the efficiency of our approach compare with baseline methods?
    \item \textbf{RQ3:} How our approach is affected by hyperparameters?
    \end{itemize}

\subsection{Experimental Setup}
\begin{table}
  \caption{Details of datasets and models}
  \label{Datasets details}
  \centering
  \scalebox{0.85}{
  \begin{tabular}{cccccc}
    \hline
    Dataset  & Classes  & Features & Training Set & Test Set  & Model    \\
    \hline
    Purchase-20       & 20  & 600  & 38758 & 9689 & FC(256)\\
    HAR       & 6  & 561  & 8238 & 2060 & FC(256)\\
    MNIST             & 10  & 28$\times$28  & 60000  & 10000  & FC(256$\times$256)\\
    CIFAR-10     & 10  & 32$\times$32$\times$3  & 50000  & 10000  & VGG16\\
    \hline
  \end{tabular}
  }
\end{table}
\emph{Datasets and Models. }We conduct experiments on four popular classification datasets in machine unlearning research: Purchase-100 \cite{shokri2017membership}, Human Activity Recognition (HAR) \cite{bulbul2018human}, MNIST \cite{lecun1998gradient}, and CIFAR-10\cite{krizhevsky2009learning}. 
Details of the dataset and models are in Table \ref{Datasets details}. 
Considering the application scenario, the data points in $\mathcal{D}_U$ are randomly selected from the training set. The confusion labels are randomly selected. The parameters for models training are described in \cite{anony2023code}.

\emph{Baselines. }We compare our approach with four widely used unlearning mechanisms: 1) Retraining, as the naive unlearning method, trains the model from scratch on $\mathcal{D} \backslash \mathcal{D}_U$. 
2) SISA \cite{bourtoule2021machine} shards the training data and then trains them separately. It yields predictions obtained by submodel majority voting. 
3) AML \cite{graves2021amnesiac} records the gradient data of each batch during the training phase. If unlearning is to be performed, the gradients of the affected batches are directly removed. 4) FU \cite{warnecke2021machine} performs closed updates of model parameters based on influence functions to forget specific training data labels or features. Considering the impact on model performance, we use the second-order update method for FU.

\subsection{Utility Guarantee}
\label{Utility Guarantee}
We evaluate each unlearning method using three metrics that reflect both effectiveness and side effects: 1) the change in accuracy on the test set $\Delta A_{tes}$, 2) the change in accuracy on the retained data not requested for unlearning $\Delta A_{res}$, and 3) the accuracy drop on the unlearned data, computed as $\Delta A_u = A_{u_b} - A_{u_a}$, reflects the extent of forgetting, where $A_{u_b}$ and $A_{u_a}$ represent the model’s accuracy on the unlearned data before and after unlearning, respectively. A larger $\Delta A_{u}$ indicates a more effective forgetting process, addressing the data holder’s concern. Meanwhile, smaller $\Delta A_{tes}$ and $\Delta A_{res}$ suggest lower negative impact on the model’s generalization and retained knowledge. In the case of single-point unlearning, PRUNE can achieve complete forgetting, as theoretically guaranteed in Theorem~\ref{the:main}.

To evaluate unlearning multiple data points, Table~\ref{effectiveness} reports model accuracy under different unlearning algorithms, with $\mathcal{D}_U$ sizes ranging from 100 to 500, sampled randomly from training batches. For methods like retraining and SISA, predictions on $\mathcal{D}_U$ barely change after unlearning, and FU shows only a slight drop—reflecting the model’s generalization ability. However, this offers little assurance to data holders or auditors that unlearning has occurred.
In contrast, both AML and PRUNE yield a significant drop (over 90\%) in accuracy on $\mathcal{D}_U$, effectively “confusing” the model on forgotten data. This allows auditors to verify unlearning via model outputs, without inspecting parameters. Notably, AML incurs considerable performance loss on the remaining data. For example, it suffers from 12.59\% drops on the accuracy of the test set for MNIST after unlearning 500 samples. PRUNE, however, confines changes to $\mathcal{D}U$, keeping $A{tes}$ nearly unchanged on Purchase, HAR, and MNIST, and within 4\% on CIFAR-10, demonstrating strong performance preservation.

\begin{table*}[t]
\belowrulesep=0pt
\aboverulesep=0pt
  \caption{Performance comparison when different numbers of data points are unlearned.}
  \label{effectiveness}
  \centering
  \scalebox{0.90}{
  {\renewcommand{\arraystretch}{1.10}
 \begin{tabular}{c|clccccc|c|clccccc}
	\toprule
Dataset & $\mathcal{D}_U$ & \multicolumn{1}{c}{Acc(\%)} & Retrain & SISA & AML & FU & PRUNE & Dataset & $\mathcal{D}_U$ & \multicolumn{1}{c}{Acc(\%)} & Retrain & SISA & AML & FU & PRUNE \\ \hline
\multirow{14}{*}{Purchase-20} & \multirow{2}{*}{0} & $A_{tes}$ & 95.08 & 90.32 & 95.16 & 92.76 & 95.30 & \multirow{14}{*}{HAR} & \multirow{2}{*}{0} & $A_{tes}$ & 95.05 & 95.29 & 94.98 & 97.78 & 95.82 \\
 &  & $A_{res}$ & 98.37 & 91.39 & 99.44 & 99.56 & 98.38 &  &  & $A_{res}$ & 96.34 & 95.89 & 97.74 & 98.79 & 97.57 \\ \cline{2-8} \cline{10-16} 
 & \multirow{3}{*}{100} & $\Delta A_{tes}$ & \textbf{-0.09} & 0.05 & 3.51 & 12.07 & {\ul 0.00} &  & \multirow{3}{*}{100} & $\Delta A_{tes}$ & \textbf{-0.04} & 1.42 & 14.26 & 0.68 & {\ul 0.00} \\
 &  & $\Delta A_{res}$ & \textbf{-0.12} & {\ul -0.02} & 2.02 & 17.99 & 0.25 &  &  & $\Delta A_{res}$ & \textbf{-0.07} & 1.26 & 0.77 & {\ul 0.33} & 1.10 \\
 &  & $\Delta A_u$ & 0.00 & 3.00 & {\ul 95.00} & 16.00 & \textbf{97.17} &  &  & $\Delta A_u$ & 0.00 & 1.00 & \textbf{100.00} & 0.00 & {\ul 92.67} \\ \cline{2-8} \cline{10-16} 
 & \multirow{3}{*}{200} & $\Delta A_{tes}$ & 0.11 & \textbf{-0.36} & 3.94 & 11.29 & {\ul 0.00} &  & \multirow{3}{*}{200} & $\Delta A_{tes}$ & \textbf{0.13} & 2.02 & 14.75 & 24.88 & {\ul 0.21} \\
 &  & $\Delta A_{res}$ & {\ul -0.08} & \textbf{-0.12} & 2.56 & 17.60 & 0.49 &  &  & $\Delta A_{res}$ & \textbf{-0.02} & 1.85 & 1.36 & 24.27 & {\ul 1.34} \\
 &  & $\Delta A_u$ & 4.00 & 1.50 & \textbf{100.00} & 14.50 & {\ul 97.70} &  &  & $\Delta A_u$ & 1.00 & 2.50 & \textbf{99.00} & 18.50 & {\ul 91.17} \\ \cline{2-8} \cline{10-16} 
 & \multirow{3}{*}{300} & $\Delta A_{tes}$ & {\ul -0.09} & \textbf{-0.35} & 4.17 & 13.52 & 0.00 &  & \multirow{3}{*}{300} & $\Delta A_{tes}$ & \textbf{0.15} & 4.27 & 14.46 & 16.84 & {\ul 0.32} \\
 &  & $\Delta A_{res}$ & {\ul -0.10} & \textbf{-0.34} & 0.96 & 19.40 & 0.74 &  &  & $\Delta A_{res}$ & \textbf{0.19} & 3.84 & {\ul 1.01} & 17.56 & 2.34 \\
 &  & $\Delta A_u$ & 1.67 & 1.66 & \textbf{99.33} & 8.34 & {\ul 95.00} &  &  & $\Delta A_u$ & 1.33 & 5.34 & \textbf{97.33} & 16.33 & {\ul 92.45} \\ \cline{2-8} \cline{10-16} 
 & \multirow{3}{*}{500} & $\Delta A_{tes}$ & {\ul 0.01} & 0.22 & 4.66 & 10.02 & \textbf{0.00} &  & \multirow{3}{*}{500} & $\Delta A_{tes}$ & \textbf{0.03} & 7.30 & 14.22 & 23.79 & {\ul 0.71} \\
 &  & $\Delta A_{res}$ & \textbf{-0.08} & 0.07 & 3.14 & 16.93 & {\ul 1.23} &  &  & $\Delta A_{res}$ & \textbf{0.16} & 8.12 & {\ul 0.72} & 25.25 & 3.50 \\
 &  & $\Delta A_u$ & 3.60 & 1.20 & \textbf{100.00} & 10.00 & {\ul 95.49} &  &  & $\Delta A_u$ & 0.20 & 7.00 & \textbf{95.20} & 27.00 & {\ul 93.60} \\ \hline
\multirow{14}{*}{MNIST} & \multirow{2}{*}{0} & $A_{tes}$ & 98.04 & 96.05 & 96.37 & 98.43 & 98.00 & \multirow{14}{*}{Cifar-10} & \multirow{2}{*}{0} & $A_{tes}$ & 87.64 & 86.66 & 86.73 & 86.36 & 87.63 \\
 &  & $A_{res}$ & 99.89 & 96.14 & 99.11 & 99.90 & 99.64 &  &  & $A_{res}$ & 97.81 & 97.29 & 96.03 & 99.45 & 97.46 \\ \cline{2-8} \cline{10-16} 
 & \multirow{3}{*}{100} & $\Delta A_{tes}$ & {\ul -0.13} & 0.04 & 9.53 & \textbf{-0.15} & 0.00 &  & \multirow{3}{*}{100} & $\Delta A_{tes}$ & \textbf{-0.49} & {\ul 0.05} & 11.59 & 2.64 & 1.04 \\
 &  & $\Delta A_{res}$ & 0.00 & {\ul -0.01} & 1.16 & \textbf{-0.04} & 0.16 &  &  & $\Delta A_{res}$ & \textbf{-0.13} & {\ul 0.13} & 5.71 & 13.37 & 1.45 \\
 &  & $\Delta A_u$ & 1.00 & 1.00 & {\ul 95.00} & 0.00 & \textbf{96.60} &  &  & $\Delta A_u$ & 10.00 & 0.00 & \textbf{95.00} & 8.74 & {\ul 92.20} \\ \cline{2-8} \cline{10-16} 
 & \multirow{3}{*}{200} & $\Delta A_{tes}$ & \textbf{-0.10} & {\ul -0.05} & 9.85 & 5.07 & 0.00 &  & \multirow{3}{*}{200} & $\Delta A_{tes}$ & {\ul 0.72} & \textbf{0.01} & 13.16 & 2.85 & 2.30 \\
 &  & $\Delta A_{res}$ & {\ul 0.00} & \textbf{-0.02} & 0.74 & 6.99 & 0.32 &  &  & $\Delta A_{res}$ & {\ul 1.10} & \textbf{0.12} & 8.60 & 13.95 & 2.81 \\
 &  & $\Delta A_u$ & 2.50 & 0.50 & {\ul 94.00} & 8.00 & \textbf{95.57} &  &  & $\Delta A_u$ & 7.50 & 1.30 & \textbf{93.38} & 9.69 & {\ul 90.30} \\ \cline{2-8} \cline{10-16} 
 & \multirow{3}{*}{300} & $\Delta A_{tes}$ & \textbf{-0.19} & {\ul -0.05} & 12.22 & 7.13 & 0.00 &  & \multirow{3}{*}{300} & $\Delta A_{tes}$ & {\ul 0.48} & \textbf{-0.18} & 11.09 & 2.13 & 3.01 \\
 &  & $\Delta A_{res}$ & {\ul 0.00} & \textbf{-0.01} & 0.20 & 8.30 & 0.49 &  &  & $\Delta A_{res}$ & {\ul 1.05} & \textbf{-0.07} & 6.66 & 14.11 & 3.94 \\
 &  & $\Delta A_u$ & 1.33 & 0.33 & {\ul 97.67} & 7.34 & \textbf{98.00} &  &  & $\Delta A_u$ & 9.67 & 0.22 & \textbf{95.50} & 12.59 & {\ul 89.83} \\ \cline{2-8} \cline{10-16} 
 & \multirow{3}{*}{500} & $\Delta A_{tes}$ & \textbf{-0.02} & 0.07 & 12.59 & 0.72 & {\ul 0.00} &  & \multirow{3}{*}{500} & $\Delta A_{tes}$ & {\ul 0.13} & \textbf{0.08} & 12.11 & 5.24 & 3.98 \\
 &  & $\Delta A_{res}$ & {\ul 0.00} & \textbf{-0.01} & 2.12 & 2.44 & 0.80 &  &  & $\Delta A_{res}$ & {\ul 0.40} & \textbf{-0.05} & 5.52 & 16.93 & 5.63 \\
 &  & $\Delta A_u$ & 1.60 & 0.40 & \textbf{97.00} & 2.20 & {\ul 96.43} &  &  & $\Delta A_u$ & 11.20 & 0.80 & \textbf{96.85} & 14.00 & {\ul 93.20} \\ \bottomrule
\end{tabular}
}
}
\end{table*}

\subsection{Efficiency Comparison}
\begin{figure*} []
	\centering  
        \subfloat[Purchase-20]{\includegraphics[width=0.25\linewidth]{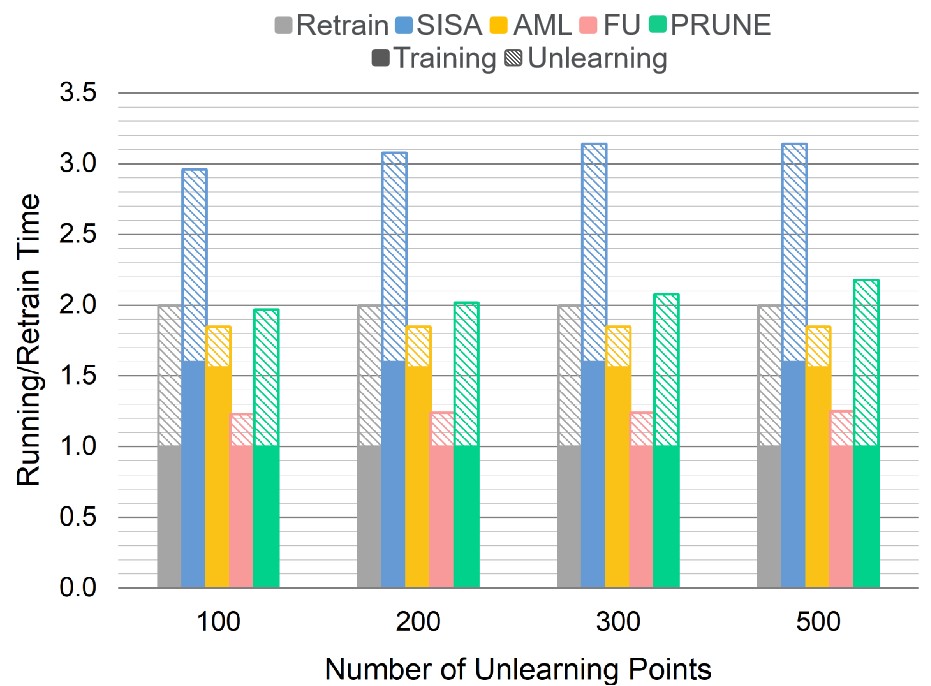}%
        \label{sub.1}}
        \subfloat[HAR]{\includegraphics[width=0.25\linewidth]{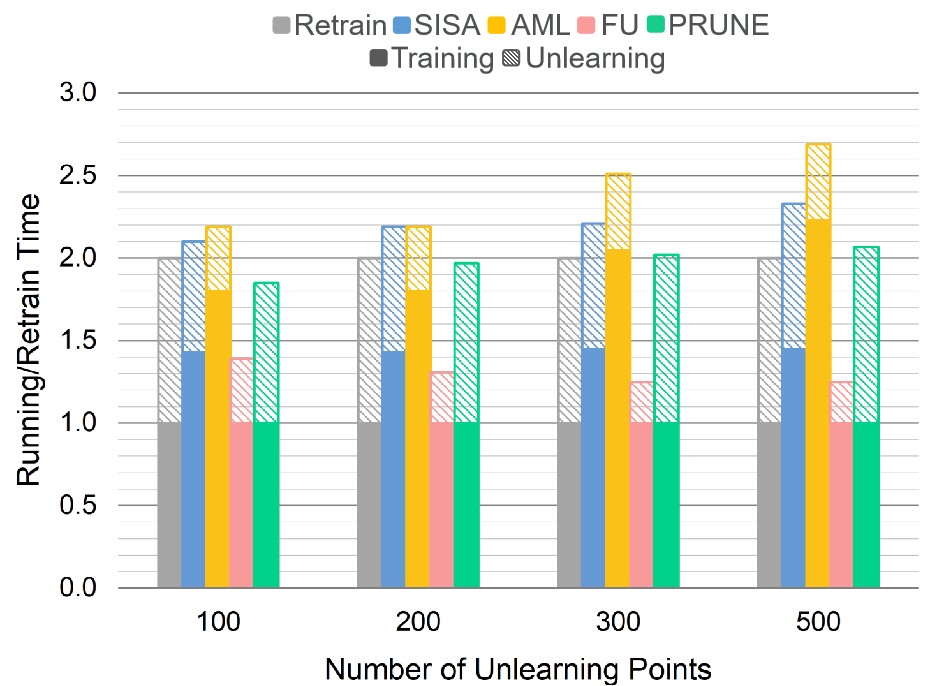}%
        \label{sub.2}}
        \subfloat[MNIST]{\includegraphics[width=0.25\linewidth]{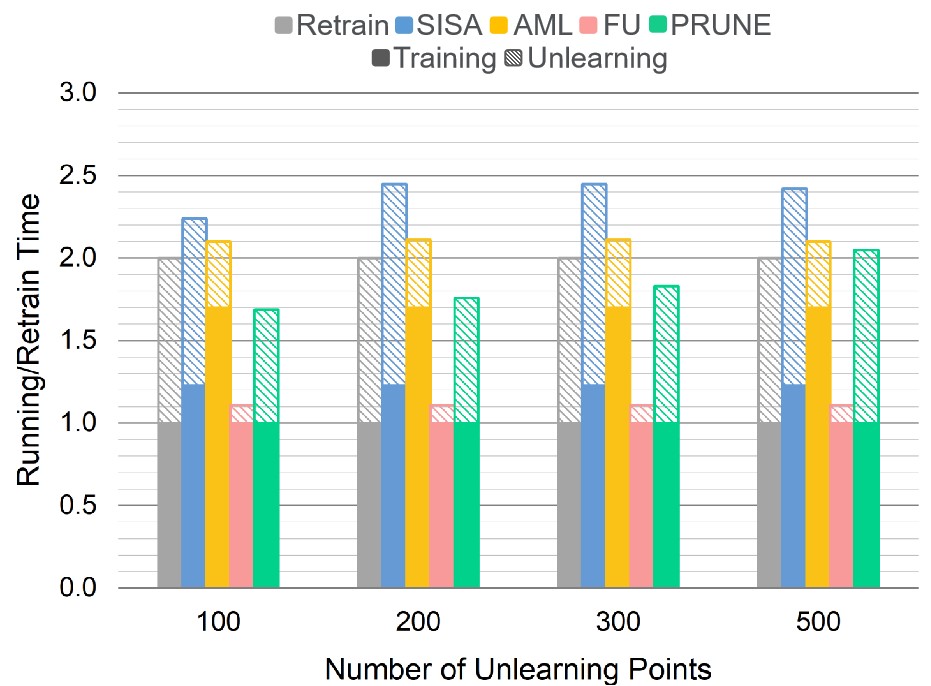}%
        \label{sub.3}}
        \subfloat[CIFAR-10]{\includegraphics[width=0.25\linewidth]{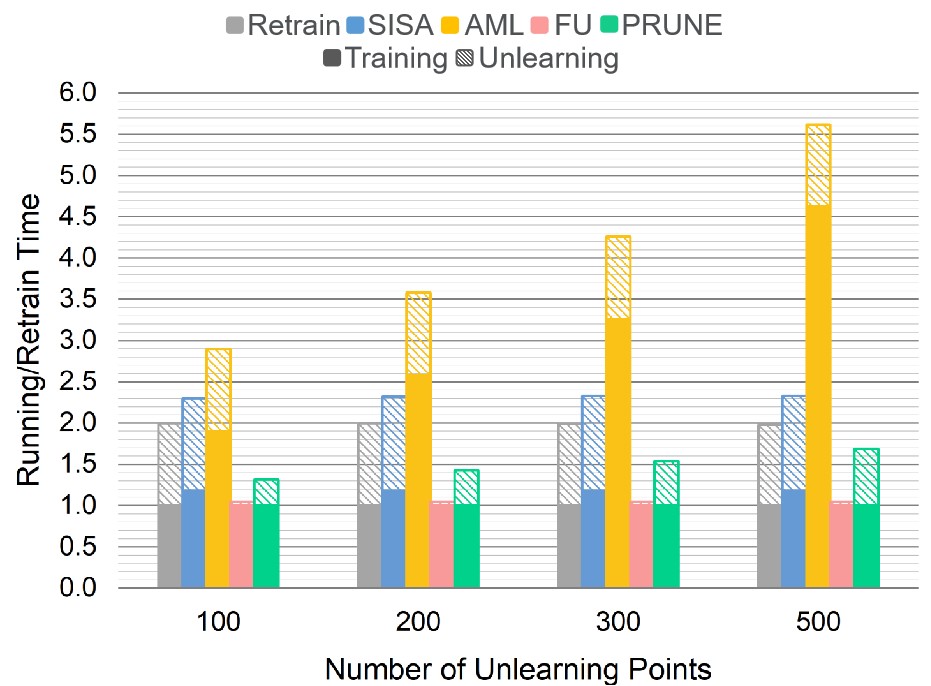}%
        \label{sub.5}}

	\caption{The efficiency of different methods to unlearn on multiple datasets. }
	\label{efficiency}
\end{figure*}

We evaluate the efficiency of different unlearning methods by recording the total time for training and unlearning, normalized by the training-from-scratch time (i.e., Retrain). As shown in Figure~\ref{efficiency}, PRUNE and FU match Retrain in training time, while AML and SISA are significantly slower, especially AML, whose cost increases with more unlearning data due to batch-wise parameter updates. AML also requires training-time instrumentation and may introduce privacy risks. In unlearning time, FU is the fastest owing to its closed-form update. PRUNE performs comparably to AML and is more stable, particularly on complex models like VGG16, where it only modifies the fully connected layers. Overall, PRUNE offers a strong balance between efficiency and applicability to real-world, large-scale models.

\subsection{Hyper-parameter Tuning}
\subsubsection*{Varying the number of centroids $K$}
\begin{figure*}[] 
	\centering  
        \subfloat[Purchase-20]{\includegraphics[width=0.24\linewidth]{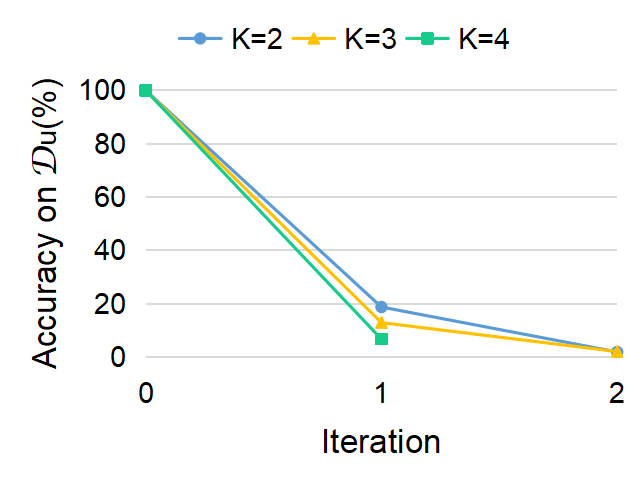}%
        \label{sub1.1}}
        \subfloat[HAR]{\includegraphics[width=0.24\linewidth]{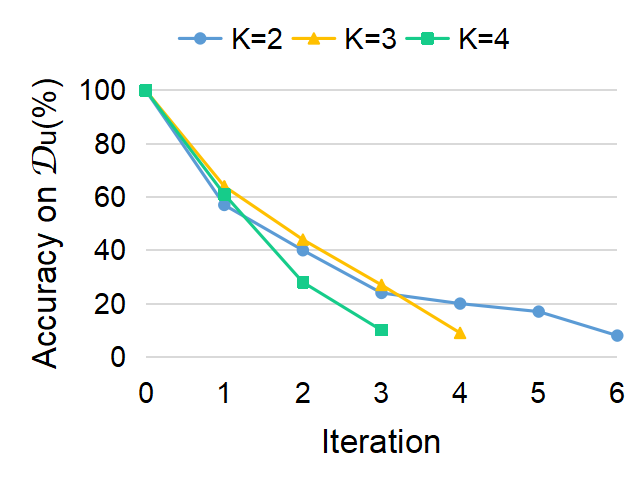}%
        \label{sub1.2}}
        \subfloat[MNIST]{\includegraphics[width=0.24\linewidth]{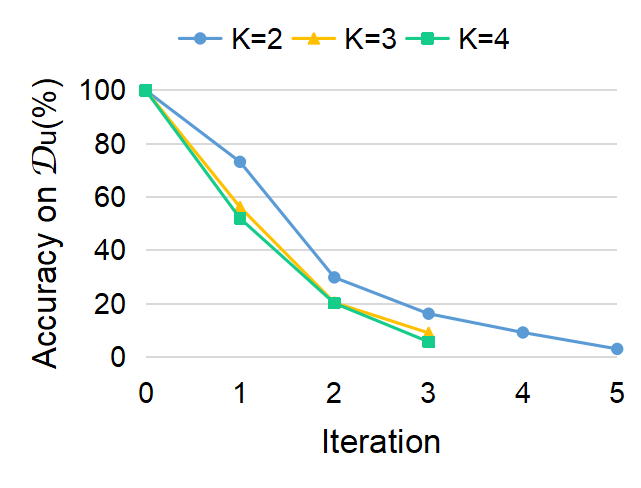}%
        \label{sub1.3}}
        \subfloat[CIFAR-10]{\includegraphics[width=0.24\linewidth]{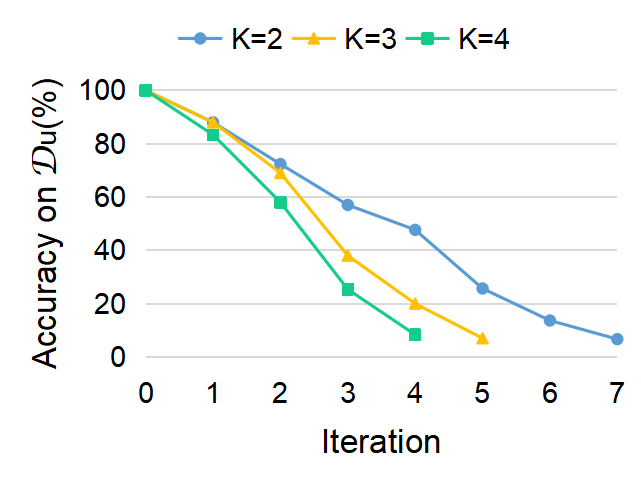}%
        \label{sub1.5}}
        
	\caption{The convergence of PRUNE w.r.t the number of centroids $K$ ($|\mathcal{D}_U|=100$).}
	\label{parameter}
\end{figure*}

Before executing our algorithm, it is necessary to determine the number of centroids $K$. That is, the number of representative data points chosen by clustering. $K$ impacts both the quality of the generated patch network and the number of optimized $m(x)$ per iteration, thus affecting convergence and time complexity. As shown in Figure~\ref{parameter}, increasing $K$ reduces the required number of iterations, but raises per-iteration computational cost. Ignoring other factors, the overall time complexity is approximately $\mathcal{O}(K \times \text{Iteration})$. For image datasets, we find that $K=2$ offers the lowest cost, as fewer centroids can represent larger clusters and influence more data points through each $m(x)$.

\subsubsection*{Varying the scale of $\mathcal{D}_U$}
\begin{figure}[]
  \centering
        \subfloat[HAR]{\includegraphics[width=0.45\linewidth]{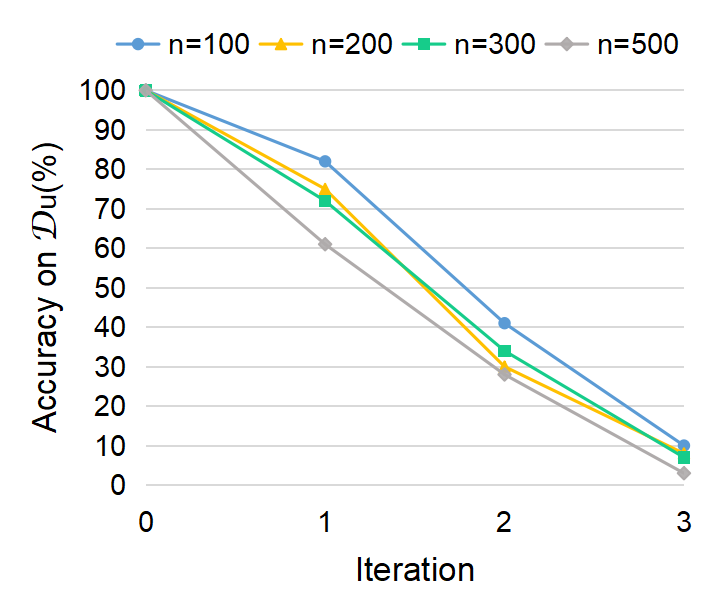}%
        \label{sub2.1}}
        \subfloat[MNIST]{\includegraphics[width=0.45\linewidth]{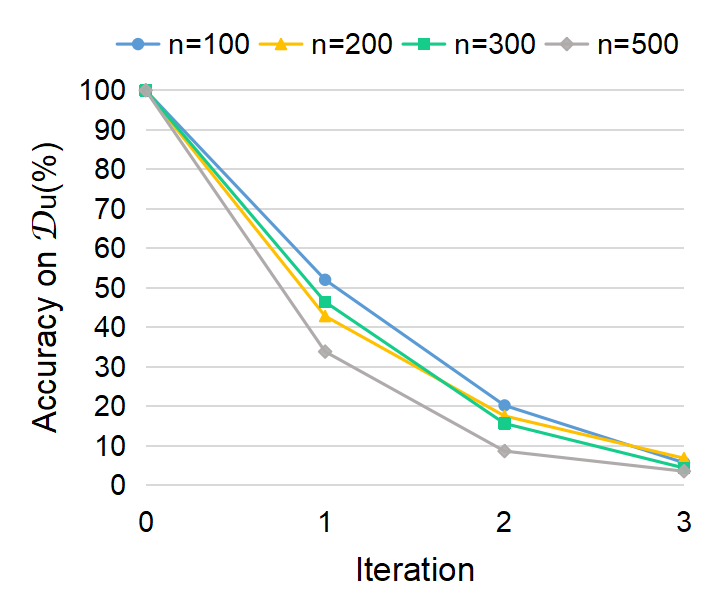}%
        \label{sub2.2}}
        
  \caption{The convergence of PRUNE w.r.t the scale of $\mathcal{D}_U$ ($K=4$).}
  \label{fig4}
\end{figure}

We evaluate convergence behavior on the HAR and MNIST datasets, with results shown in Figure~\ref{fig4}. Other datasets' in \cite{anony2023code}. Despite varying iteration counts, all settings yield consistent conclusions.
We can observe that PRUNE converges quickly no matter how the scale of $\mathcal{D}_U$ varies. This proves the superiority of our algorithm on multipoint withdrawal. However, as the size of $\mathcal{D}_U$ increases, there is a tendency for PRUNE to converge faster. This trend is most obvious on the HAR dataset. When $|\mathcal{D}_U|=500$, the average accuracy of the model on $\mathcal{D}_U$ after 3 iterations is 3.1\%. It is $1/3$ of the accuracy of the model when $|\mathcal{D}_U|=100$. 

\section{Conclusion}
\label{Conclusion}
In this paper, we propose a novel certifiable unlearning algorithm PRUNE to erase specific data by adding a targeted patch network to the original model. Unlike existing works, our approach eases data holders' concern by providing easily measurable forgetting effect (examining the model's prediction on the unlearned data) while not affecting the model's overall performance.
Extensive experiments have demonstrated that PRUNE can efficiently unlearn multiple data points. 
In order to support further development of certifiable machine unlearning, we make all the codes available at the public repository \cite{anony2023code}.

\section*{Acknowledgment}

This research is supported by the State Key Laboratory of Industrial
Control Technology, China (Grant No. ICT2024C01).

{\footnotesize \bibliographystyle{IEEEtran}
\bibliography{ref}}

\begin{thebibliography}{10}
\providecommand{\url}[1]{#1}
\csname url@samestyle\endcsname
\providecommand{\newblock}{\relax}
\providecommand{\bibinfo}[2]{#2}
\providecommand{\BIBentrySTDinterwordspacing}{\spaceskip=0pt\relax}
\providecommand{\BIBentryALTinterwordstretchfactor}{4}
\providecommand{\BIBentryALTinterwordspacing}{\spaceskip=\fontdimen2\font plus
\BIBentryALTinterwordstretchfactor\fontdimen3\font minus \fontdimen4\font\relax}
\providecommand{\BIBforeignlanguage}[2]{{%
\expandafter\ifx\csname l@#1\endcsname\relax
\typeout{** WARNING: IEEEtran.bst: No hyphenation pattern has been}%
\typeout{** loaded for the language `#1'. Using the pattern for}%
\typeout{** the default language instead.}%
\else
\language=\csname l@#1\endcsname
\fi
#2}}
\providecommand{\BIBdecl}{\relax}
\BIBdecl

\bibitem{regulation2018general}
G.~D.~P. Regulation, ``General data protection regulation (gdpr),'' \emph{Intersoft Consulting, Accessed in October}, vol.~24, no.~1, 2018.

\bibitem{ginart2019making}
A.~A. Ginart, M.~Y. Guan, G.~Valiant, and J.~Zou, ``Making ai forget you: data deletion in machine learning,'' in \emph{Proceedings of the 33rd International Conference on Neural Information Processing Systems}, 2019, pp. 3518--3531.

\bibitem{bourtoule2021machine}
L.~Bourtoule, V.~Chandrasekaran, C.~A. Choquette-Choo, H.~Jia, A.~Travers, B.~Zhang, D.~Lie, and N.~Papernot, ``Machine unlearning,'' in \emph{2021 IEEE Symposium on Security and Privacy (SP)}.\hskip 1em plus 0.5em minus 0.4em\relax IEEE, 2021, pp. 141--159.

\bibitem{zhou2025decoupled}
Y.~Zhou, D.~Zheng, Q.~Mo, R.~Lu, K.-Y. Lin, and W.-S. Zheng, ``Decoupled distillation to erase: A general unlearning method for any class-centric tasks,'' in \emph{Proceedings of the Computer Vision and Pattern Recognition Conference}, 2025, pp. 20\,350--20\,359.

\bibitem{yan2022arcane}
H.~Yan, X.~Li, Z.~Guo, H.~Li, F.~Li, and X.~Lin, ``Arcane: An efficient architecture for exact machine unlearning,'' in \emph{Proceedings of the Thirty-First International Joint Conference on Artificial Intelligence, IJCAI-22}, 2022, pp. 4006--4013.

\bibitem{foster2024fast}
J.~Foster, S.~Schoepf, and A.~Brintrup, ``Fast machine unlearning without retraining through selective synaptic dampening,'' in \emph{Proceedings of the AAAI conference on artificial intelligence}, vol.~38, no.~11, 2024, pp. 12\,043--12\,051.

\bibitem{hu2024separate}
X.~Hu, D.~Li, B.~Hu, Z.~Zheng, Z.~Liu, and M.~Zhang, ``Separate the wheat from the chaff: Model deficiency unlearning via parameter-efficient module operation,'' in \emph{Proceedings of the AAAI Conference on Artificial Intelligence}, vol.~38, no.~16, 2024, pp. 18\,252--18\,260.

\bibitem{zhang2022prompt}
Z.~Zhang, Y.~Zhou, X.~Zhao, T.~Che, and L.~Lyu, ``Prompt certified machine unlearning with randomized gradient smoothing and quantization,'' \emph{Advances in Neural Information Processing Systems}, vol.~35, pp. 13\,433--13\,455, 2022.

\bibitem{chien2023efficient}
E.~Chien, C.~Pan, and O.~Milenkovic, ``Efficient model updates for approximate unlearning of graph-structured data,'' in \emph{International Conference on Learning Representations}, 2023.

\bibitem{wu2022puma}
G.~Wu, M.~Hashemi, and C.~Srinivasa, ``Puma: Performance unchanged model augmentation for training data removal,'' in \emph{Proceedings of the AAAI Conference on Artificial Intelligence}, vol.~36, no.~8, 2022, pp. 8675--8682.

\bibitem{wu2023gif}
J.~Wu, Y.~Yang, Y.~Qian, Y.~Sui, X.~Wang, and X.~He, ``Gif: A general graph unlearning strategy via influence function,'' in \emph{Proceedings of the ACM Web Conference 2023}, 2023, pp. 651--661.

\bibitem{dong2021towards}
G.~Dong, J.~Sun, X.~Wang, X.~Wang, and T.~Dai, ``Towards repairing neural networks correctly,'' in \emph{2021 IEEE 21st International Conference on Software Quality, Reliability and Security (QRS)}.\hskip 1em plus 0.5em minus 0.4em\relax IEEE, 2021, pp. 714--725.

\bibitem{ma2024vere}
J.~Ma, P.~Yang, J.~Wang, Y.~Sun, C.-C. Huang, and Z.~Wang, ``Vere: Verification guided synthesis for repairing deep neural networks,'' in \emph{Proceedings of the 46th IEEE/ACM International Conference on Software Engineering}, 2024, pp. 1--13.

\bibitem{sohn2022arachne}
J.~Sohn, S.~Kang, and S.~Yoo, ``Arachne: Search based repair of deep neural networks,'' \emph{ACM Transactions on Software Engineering and Methodology}, 2022.

\bibitem{anony2023code}
Anonymous, ``The repository of code and data to support patching based repair framework for certifiable unlearning,'' \url{https://github.com/dummyPRUNE/PRUNE2024}, 2024.

\bibitem{lee2019towards}
G.-H. Lee, D.~Alvarez-Melis, and T.~S. Jaakkola, ``Towards robust, locally linear deep networks,'' in \emph{International Conference on Learning Representations}, 2019.

\bibitem{fu2022sound}
F.~Fu and W.~Li, ``Sound and complete neural network repair with minimality and locality guarantees,'' in \emph{International Conference on Learning Representations}, 2022.

\bibitem{shokri2017membership}
R.~Shokri, M.~Stronati, C.~Song, and V.~Shmatikov, ``Membership inference attacks against machine learning models,'' in \emph{2017 IEEE symposium on security and privacy (SP)}.\hskip 1em plus 0.5em minus 0.4em\relax IEEE, 2017, pp. 3--18.

\bibitem{bulbul2018human}
E.~Bulbul, A.~Cetin, and I.~A. Dogru, ``Human activity recognition using smartphones,'' in \emph{2018 2nd international symposium on multidisciplinary studies and innovative technologies (ismsit)}.\hskip 1em plus 0.5em minus 0.4em\relax IEEE, 2018, pp. 1--6.

\bibitem{lecun1998gradient}
Y.~LeCun, L.~Bottou, Y.~Bengio, and P.~Haffner, ``Gradient-based learning applied to document recognition,'' \emph{Proceedings of the IEEE}, vol.~86, no.~11, pp. 2278--2324, 1998.

\bibitem{krizhevsky2009learning}
A.~Krizhevsky \emph{et~al.}, ``Learning multiple layers of features from tiny images,'' \emph{https://www. cs. toronto. edu/kriz/learning-features-2009-TR. pdf}, 2009.

\bibitem{graves2021amnesiac}
L.~Graves, V.~Nagisetty, and V.~Ganesh, ``Amnesiac machine learning,'' in \emph{Proceedings of the AAAI Conference on Artificial Intelligence}, vol.~35, no.~13, 2021, pp. 11\,516--11\,524.

\bibitem{warnecke2021machine}
A.~Warnecke, L.~Pirch, C.~Wressnegger, and K.~Rieck, ``Machine unlearning of features and labels,'' in \emph{Proc. of the 30th Network and Distributed System Security (NDSS)}, 2023.

\end{thebibliography}
\end{document}